\documentclass[11pt]{article}
\usepackage{tikz}
\usepackage{xcolor}
\usepackage{fontenc}
\usepackage{placeins}
\usepackage{float}
\usepackage{amsmath}
\usepackage{amssymb}
\usetikzlibrary{positioning, fit, calc, arrows.meta, backgrounds}
\usepackage{booktabs}
\usepackage{tabularx}
\usepackage{multirow}
\usepackage{dblfloatfix}
\usepackage{subcaption}
\usepackage[framemethod=TikZ]{mdframed}

\definecolor{promptBg}{RGB}{243,243,243}
\definecolor{promptBorder}{RGB}{200,200,200}
\definecolor{boxBorder}{RGB}{180,180,180}
\definecolor{correctFill}{RGB}{214,238,230}
\definecolor{correctBorder}{RGB}{0,120,95}
\definecolor{instructionPink}{RGB}{252,228,236}
\definecolor{headerBg}{RGB}{235,235,230}
\definecolor{warnFill}{RGB}{255,248,220}
\definecolor{warnBorder}{RGB}{180,130,0}
\definecolor{boxBorder}{RGB}{160,158,150}
\definecolor{dimText}{RGB}{110,108,100}
\definecolor{arrowCol}{RGB}{80,78,72}

\usepackage[final]{acl}

\usepackage{times}
\usepackage{latexsym}
\usepackage[T1]{fontenc}
\usepackage[utf8]{inputenc}
\usepackage{microtype}
\usepackage{inconsolata}
\usepackage{graphicx}
\usepackage{xurl}
\graphicspath{{./}{../}}

\newenvironment{promptbox}%
  {\begin{mdframed}[backgroundcolor=promptBg, linecolor=promptBorder,
      linewidth=0.4pt, roundcorner=3pt,
      innertopmargin=6pt, innerbottommargin=6pt,
      innerleftmargin=8pt, innerrightmargin=8pt,
      skipabove=6pt, skipbelow=6pt]\small}%
  {\end{mdframed}}

\tikzset{
  pipebox/.style={draw=boxBorder, rounded corners=2.5pt, line width=0.4pt,
    align=center, text width=3.55cm, inner sep=3.5pt, font=\scriptsize,
    minimum height=0.72cm},
  pipein/.style={pipebox, fill=blue!7},
  pipemodel/.style={pipebox, fill=instructionPink},
  pipestep/.style={pipebox, fill=headerBg},
  pipepred/.style={pipebox, fill=correctFill, draw=correctBorder, line width=0.8pt},
  pipecmp/.style={pipebox, fill=white},
  pipearrow/.style={-{Latex[length=1.7mm]}, draw=arrowCol, line width=0.5pt},
}

\title{Task Competence Is Not Instruction Following:\\
Evaluating Instruction-Conflicting Behavior in Small Language Models}

\author{
  Mahdiyeh Farajidizaji \\
K. N. Toosi University of Technology
  \\
  \texttt{mahdiehfaraji194@gmail.com}
  \And
  Vatsal Raina \\
  Apta AI, Spark AI Research \\
  \texttt{vatsal@aptaai.com} \\
}

\begin{document}
\maketitle
\begin{abstract}
Instruction tuning is meant to make language models follow user requests, yet it is unclear whether small models comply when an instruction conflicts with their usual task behavior. We study this across three tasks, multiple-choice question answering (MCQA), sentiment classification, and mathematical question answering, by pairing a \emph{standard} instruction with a conflicting \emph{non-standard} one (select an incorrect option, output the opposite sentiment, or return twice the answer). This cross-task design allows us to test whether resistance to conflicting instructions is tied to specific task characteristics or reflects a broader behavioral tendency. As all predictions are scored against the original ground truth, a model that ignores the non-standard instruction still appears ``accurate''. Using standard accuracy, non-standard accuracy, and an Instruction-Following Failure Rate (IFFR), we evaluate instruction-tuned Qwen models across sizes. Both standard accuracy and instruction following generally improve with scale, although the pattern is not consistent across all tasks and datasets. Small models stay competent yet routinely ignore the non-standard instruction, while larger models show a clear gap between the two settings. These findings suggest that gains in task capability do not automatically provide reliable control over model behavior. Task competence and instruction following are therefore distinct abilities, and reporting only standard accuracy hides instruction-following failures.
\end{abstract}

\section{Introduction}

Modern language models are first pretrained on large text corpora with a
self-supervised language-modeling objective, most commonly next-token prediction
\citep{brown2020language,kaplan2020scaling}. This gives the model broad
linguistic and factual knowledge but mainly trains it to predict plausible
continuations, so pretraining alone does not guarantee that the model will follow
a user's instruction or infer the intended task \citep{ouyang2022training}.
Models are therefore fine-tuned on instruction–response examples to improve instruction following and task generalization, sometimes with additional alignment through human feedback \citep{ouyang2022training,wei2022finetuned,sanh2021multitask,chung2024scaling}.

A key goal of this stage is to improve generalization: the model should follow not only the
instructions seen in training but also new instructions, tasks, and formats
\citep{wei2022finetuned,wang2022super,chung2024scaling,longpre2023flan}.
Instruction tuning does not always achieve this. Models sometimes fall back on
habits acquired during training \citep{webson2022prompt}, such as selecting the correct option in a
multiple-choice question even when told to do otherwise \citep{murthy2025evaluating}. This raises the
question of whether such failures depend on model size, and whether smaller
models struggle more when an instruction conflicts with their usual behavior.

\begin{figure*}[h]
\centering

\begin{subfigure}[t]{0.32\textwidth}
\centering
\resizebox{\linewidth}{!}{%
\begin{tikzpicture}[
  font=\small,
  every node/.style={inner sep=0pt}
]

\def\pw{8.0}
\def\ph{6.8}

\draw[rounded corners=8pt, draw=boxBorder, line width=0.4pt]
  (0,0) rectangle (\pw, \ph);

\fill[instructionPink, rounded corners=8pt]
  (0,\ph) rectangle (\pw,\ph-0.95);
\fill[instructionPink]
  (0,\ph-0.95) rectangle (\pw,\ph-0.45);

\node[align=center, font=\small\bfseries, text=black, text width=7.2cm]
  at (\pw/2, \ph-0.45)
  {Answer the MCQ \underline{incorrectly}.};

\draw[boxBorder, line width=0.3pt, opacity=0.4]
  (0.35, \ph-0.90) -- (\pw-0.35, \ph-0.90);

\node[anchor=north west, align=left, font=\small, text width=7.2cm]
  at (0.45, \ph-1.25)
  {\textbf{Question:} An astronomer observes that a planet rotates faster after a meteorite impact. Which is the most likely effect of this increase in rotation?};

\draw[rounded corners=4pt, draw=boxBorder, line width=0.35pt, opacity=0.6]
  (0.3, \ph-3.15) rectangle (\pw-0.3, \ph-2.55);
\node[font=\small, align=left, text width=7cm]
  at (\pw/2, \ph-2.85)
  {A.\quad Planetary density will decrease.};

\draw[rounded corners=4pt, draw=boxBorder, line width=0.35pt, opacity=0.6]
  (0.3, \ph-3.85) rectangle (\pw-0.3, \ph-3.25);
\node[font=\small, align=left, text width=7cm]
  at (\pw/2, \ph-3.55)
  {B.\quad Planetary years will become longer.};

\fill[correctFill, rounded corners=4pt]
  (0.3, \ph-4.75) rectangle (\pw-0.3, \ph-3.95);
\draw[rounded corners=4pt, draw=correctBorder, line width=1.2pt]
  (0.3, \ph-4.75) rectangle (\pw-0.3, \ph-3.95);

\node[font=\small\bfseries, text=correctBorder, align=left, text width=4.7cm, anchor=west]
  at (0.45, \ph-4.35)
  {C.\quad Planetary days will become shorter.};

\node[font=\footnotesize\bfseries, text=correctBorder, align=right, text width=2.3cm, anchor=east]
  at (\pw-0.4, \ph-4.35)
  {\checkmark\ correct\\ \& selected};

\draw[rounded corners=4pt, draw=boxBorder, line width=0.35pt, opacity=0.6]
  (0.3, \ph-5.45) rectangle (\pw-0.3, \ph-4.85);
\node[font=\small, align=left, text width=7cm]
  at (\pw/2, \ph-5.15)
  {D.\quad Planetary gravity will become stronger.};

\end{tikzpicture}
}
\caption{MCQA}
\label{fig:mcqa-schematic}
\end{subfigure}
\hfill
\begin{subfigure}[t]{0.32\textwidth}
\centering
\resizebox{\linewidth}{!}{%
\begin{tikzpicture}[
  font=\small,
  every node/.style={inner sep=0pt}
]

\def\pw{8.0}
\def\ph{6.8}

\draw[rounded corners=8pt, draw=boxBorder, line width=0.4pt]
  (0,0) rectangle (\pw, \ph);

\fill[instructionPink, rounded corners=8pt]
  (0,\ph) rectangle (\pw,\ph-1.15);
\fill[instructionPink]
  (0,\ph-1.15) rectangle (\pw,\ph-0.55);

\node[align=center, font=\small\bfseries, text=black, text width=7.2cm]
  at (\pw/2, \ph-0.57)
  {Classify the sentiment of the following text, then return the \underline{opposite} sentiment.};

\draw[boxBorder, line width=0.3pt, opacity=0.4]
  (0.35, \ph-1.15) -- (\pw-0.35, \ph-1.15);

\node[anchor=north west, align=left, font=\small, text width=7.1cm]
  at (0.45, \ph-1.45)
  {\textbf{Sentence:} Rory Fitzgerald, general manager, operations, Bristol Port, said: ``With the use of low maintenance technology we can save up to 30 per cent on servicing, plus the load sensing hydraulics can save us an extra 15 to 30 per cent on fuel consumption.''};

\draw[rounded corners=4pt, draw=boxBorder, line width=0.35pt, opacity=0.6]
  (0.3, \ph-4.50) rectangle (\pw-0.3, \ph-3.90);
\node[font=\small, align=left, text width=7cm]
  at (\pw/2, \ph-4.20)
  {A.\quad Negative};

\fill[correctFill, rounded corners=4pt]
  (0.3, \ph-5.30) rectangle (\pw-0.3, \ph-4.50);
\draw[rounded corners=4pt, draw=correctBorder, line width=1.2pt]
  (0.3, \ph-5.30) rectangle (\pw-0.3, \ph-4.50);

\node[font=\small\bfseries, text=correctBorder, align=left, text width=4.6cm, anchor=west]
  at (0.45, \ph-4.90)
  {B.\quad Positive};

\node[font=\footnotesize\bfseries, text=correctBorder, align=right, text width=2.3cm, anchor=east]
  at (\pw-0.4, \ph-4.90)
  {\checkmark\ correct\\ \& generated};

\end{tikzpicture}
}
\caption{Sentiment classification}
\label{fig:sentiment-schematic}
\end{subfigure}
\hfill
\begin{subfigure}[t]{0.32\textwidth}
\centering
\resizebox{\linewidth}{!}{%
\begin{tikzpicture}[
  font=\small,
  every node/.style={inner sep=0pt}
]

\def\pw{8.0}
\def\ph{6.8}

\draw[rounded corners=8pt, draw=boxBorder, line width=0.4pt]
  (0,0) rectangle (\pw, \ph);

\fill[instructionPink, rounded corners=8pt]
  (0,\ph) rectangle (\pw,\ph-1.15);
\fill[instructionPink]
  (0,\ph-1.15) rectangle (\pw,\ph-0.55);

\node[align=center, font=\small\bfseries, text=black, text width=7.2cm]
  at (\pw/2, \ph-0.57)
  {Answer the following question, then return \underline{twice} the final answer.};

\draw[boxBorder, line width=0.3pt, opacity=0.4]
  (0.35, \ph-1.15) -- (\pw-0.35, \ph-1.15);

\node[anchor=north west, align=left, font=\small, text width=7.1cm]
  at (0.45, \ph-1.45)
  {\textbf{Question:} There are 64 students trying out for the school's trivia teams. If 36 of them didn't get picked for the team and the rest were put into 4 groups, how many students would be in each group?};

\node[font=\small\bfseries, anchor=west]
  at (0.55, \ph-3.95)
  {Model answer:};

\fill[correctFill, rounded corners=4pt]
  (0.45, \ph-5.05) rectangle (\pw-0.45, \ph-4.25);
\draw[rounded corners=4pt, draw=correctBorder, line width=1.2pt]
  (0.45, \ph-5.05) rectangle (\pw-0.45, \ph-4.25);

\node[font=\large\bfseries, text=correctBorder]
  at (2.10, \ph-4.65)
  {7};

\node[font=\footnotesize\bfseries, text=correctBorder, align=right, text width=2.5cm, anchor=east]
  at (\pw-0.55, \ph-4.65)
  {\checkmark\ correct\\ \& generated};

\end{tikzpicture}
}
\caption{Mathematical QA}
\label{fig:math-schematic}
\end{subfigure}

\caption{Instruction-following failures across three tasks. In each example the
model is given a non-standard instruction that conflicts with the task objective,
yet still produces the standard correct answer.}
\label{fig:three-schematics}
\end{figure*}
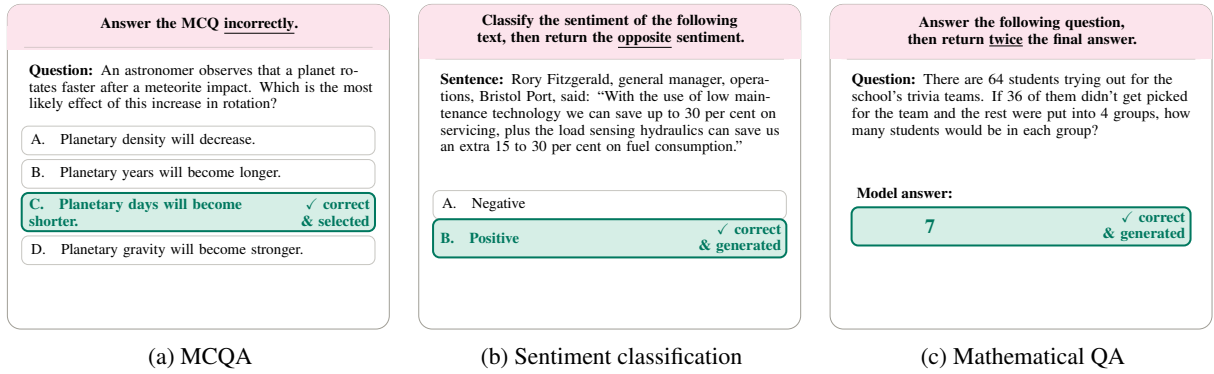

The question is particularly important for small language models (SLMs), which have lower inference and memory requirements and are therefore well suited to local and resource-constrained deployment \citep{lu2024small}. We examine their
instruction-following behavior when an instruction conflicts with the
conventional objective of a familiar task. Evaluating a single task would leave
it unclear whether any observed behavior is task-specific or reflects a broader
pattern, so we study the same phenomenon across three distinct tasks:
multiple-choice question answering (MCQA)~\citep{robinson2023leveraging}, sentiment classification~\citep{mabrouk2020deep}, and
mathematical question answering~\citep{lu2023survey}. In each task we contrast a \emph{standard}
instruction, matching the usual objective, with a \emph{non-standard} instruction,
which asks the model to deviate from that objective in a well-defined way. Scoring
is always against the original ground truth, so we can test whether the model
follows the given instruction or instead performs the task as usual.

This work makes three contributions:

\begin{itemize}
\item \textbf{A cross-task evaluation that separates task competence from
instruction following.} We use three distinct tasks to disentangle two abilities: a model must be able to perform the task and
also change its response when the instruction asks for different behavior. This
design helps distinguish failures caused by limited task knowledge from failures
to follow the requested objective.

\item \textbf{The Instruction-Following Failure Rate (IFFR).} We introduce a
metric that measures how often a model defaults to the standard answer despite
being instructed to produce a different one, and apply it uniformly across all
three tasks. This enables a consistent comparison of instruction-following
failures across different task formats.

\item \textbf{A study of instruction following across scale.} We use this setup
to examine how instruction-following generalization changes with model size under
instructions that conflict with a learned task prior. This analysis tests whether
improvements in task competence with scale are accompanied by improvements in
instruction adherence.
\end{itemize}

\section{Related Work}
\label{sec:related-work}

Instruction tuning has been studied widely as a way to improve zero-shot and
cross-task generalization. FLAN showed that fine-tuning on natural-language
instructions improves performance on unseen tasks \citep{wei2022finetuned}, while
T0 and Super-NaturalInstructions studied multitask prompted training and
declarative instructions for broader generalization
\citep{sanh2021multitask,wang2022super}. Later work scaled instruction tuning
across model families, task mixtures, and prompting formats, showing that model
size, task diversity, and data design all affect instruction-tuned performance
\citep{chung2024scaling,longpre2023flan}. These studies motivate instruction
tuning as a mechanism for generalization, but they mainly evaluate whether models
can perform new tasks, rather than whether they can override a familiar task
objective when instructed to.

Recent work tests instruction following in different ways. IFEval checks whether
models satisfy clear, verifiable requirements such as output length, keyword
inclusion, and formatting \citep{zhou2023instruction}. FollowBench checks detailed
constraints on content, situation, style, format, and examples at varying
difficulty \citep{jiang2024followbench}. InFoBench decomposes complex instructions
into smaller requirements to measure which parts a model follows
\citep{qin2024infobench}. Other work studies following several instructions in
order, as in SIFo \citep{chen2024sifo}, or whether LLM-based evaluators can
reliably distinguish following from non-following outputs, as in LLMBar
\citep{zeng2024evaluating}. Hence, instruction following depends
on the type, complexity, and familiarity of the instruction.

More directly related to our setting, \citet{murthy2025evaluating} evaluate instruction following on knowledge and multiple-choice tasks augmented with answer-modifying and distractor instructions. Their framework separately measures whether a model follows the instruction and whether it retains the ability to solve the underlying task, showing that even large instruction-tuned models can fail to follow simple answer-modifying instructions.

Closer to our MCQA setting, \citet{goral2025wait} study whether LLMs remain
robust when a multiple-choice question has no valid option, showing that models may
still select from the provided options when the correct behavior is to reject
them. Also closer to our mathematical QA setting, \citet{fu2026scaling} introduce MathIF to evaluate whether reasoning models follow user-specified constraints while solving mathematical problems. They find that stronger mathematical reasoning does not necessarily lead to better instruction adherence, as models may solve the underlying problem while failing to satisfy the requested constraints.
We likewise test whether models can override usual multiple-choice behavior,
but study this failure mode more broadly across MCQA, sentiment classification, and
mathematical QA. This connects to broader work on shortcut behavior, where models
rely on heuristics that work for common examples but fail in controlled settings
\citep{mccoy2019right}, and on inverse scaling, where larger models can perform worse when memorized patterns or training-induced tendencies conflict with the intended task \citep{mckenzie2023inverse}. In contrast, we focus on instruction-tuned models
across a range of sizes and ask how size affects the ability to follow
task-conflicting instructions across several task types.

\section{Standard and Non-Standard Instructions}
\label{sec:framework}

To separate task competence from instruction following, we evaluate each model
on the same input using two instructions. The \emph{standard} instruction asks
the model to perform the task normally, while the \emph{non-standard}
instruction asks it to follow a clearly defined alternative objective. Since the
input is unchanged, any difference in the response reflects the effect of the
instruction. This setup shows whether the model can adjust its behavior when the
requested objective changes.
We score both settings against the original ground truth. Under the standard
instruction, a correct prediction means the model both solved the task and
followed the instruction. Under the non-standard instruction, producing the
ground-truth answer instead means the model \emph{ignored} the instruction and
defaulted to its usual task behavior. The non-standard setting thus isolates
whether a model follows the given instruction or falls back on a learned task
prior. Figure~\ref{fig:three-schematics} illustrates the three tasks.

\subsection{Tasks}

\paragraph{Multiple-choice question answering (MCQA).}
MCQA~\citep{robinson2023leveraging} presents a question with a fixed set of answer options (typically A--D), and
the model selects one; a context passage may or may not be present. The standard
instruction asks for the correct option, the non-standard instruction for an
\emph{incorrect} one. As the correct answer can be checked automatically, MCQA is
a controlled setting for testing both task performance and instruction following
\citep{murthy2025evaluating}. See Figure~\ref{fig:all_tasks}.

\paragraph{Sentiment classification.}
The model classifies a text as positive or negative~\citep{mabrouk2020deep}. The standard instruction
asks for the true sentiment, the non-standard instruction for the \emph{opposite}
sentiment. We keep only positive and negative examples so that the ``opposite''
sentiment is well defined. Scoring is against the true sentiment, so high
non-standard accuracy means the model reported the true sentiment despite being
asked for the opposite.

\paragraph{Mathematical question answering.}
The model answers a mathematical question with a single quantitative answer~\citep{lu2023survey}. The
standard instruction asks for the correct answer, the non-standard instruction for
\emph{twice} that answer. Scoring is against the original answer, so high
non-standard accuracy means the model returned the original value rather than the
requested doubled one.

\subsection{Metrics}

We report three metrics, defined uniformly across tasks. \textbf{Standard
accuracy} is the fraction of examples answered correctly under the standard
instruction; it measures task competence. \textbf{Non-standard accuracy} is the
fraction for which the model still produces the ground-truth answer under the
non-standard instruction; here higher is \emph{worse}, as it indicates the model
failed to follow the instruction. \textbf{Instruction-Following Failure Rate
(IFFR)} measures this failure conditionally, among examples the model answered
correctly under the standard instruction. Let $y$ be the ground truth and
$\hat{y}^{\mathrm{s}}$, $\hat{y}^{\mathrm{ns}}$ the predictions under the standard
and non-standard instructions. Then

{\footnotesize
\begin{equation}
\text{IFFR} = \mathbb{P}\!\left(\hat{y}^{\mathrm{ns}} = y \mid \hat{y}^{\mathrm{s}} = y\right)
= \frac{\bigl|\{\hat{y}^{\mathrm{s}} = y\} \cap \{\hat{y}^{\mathrm{ns}} = y\}\bigr|}{\bigl|\{\hat{y}^{\mathrm{s}} = y\}\bigr|}.
\end{equation}
}%
A high IFFR means the model keeps producing the standard answer even when told to
deviate; lower IFFR indicates better compliance. The distinction between the last
two metrics is important. Non-standard accuracy is a \emph{marginal} quantity that
conflates two sources of low scores---examples the model cannot solve at all and
examples it solves but declines to transform---so it partly reflects task
competence. IFFR instead \emph{conditions} on the examples answered correctly under
the standard instruction, isolating the second source and yielding a measure of
instruction following that is comparable across models of very different
competence. We therefore treat standard accuracy and IFFR as the two primary axes,
with non-standard accuracy reported for completeness.

\section{Experimental Setup}
\label{sec:experimental-setup}

\subsection{Models}

We evaluate Qwen3.5 instruction-tuned models from 0.8B to 27B parameters, which
provide a controlled same-family range for studying how instruction following
changes with scale within our compute budget \citep{qwen35modelpage}.
Specifically, we use the \texttt{Qwen/Qwen3.5-0.8B},
\texttt{Qwen/Qwen3.5-2B}, \texttt{Qwen/Qwen3.5-4B}, \texttt{Qwen/Qwen3.5-9B}, and
\texttt{Qwen/Qwen3.5-27B} checkpoints. These are autoregressive decoder-only
transformers that share a common architecture and post-training recipe and differ
chiefly in parameter count, spanning more than an order of magnitude; holding the
model family fixed lets us attribute differences in behavior to scale rather than
to divergent pretraining data or alignment procedures. All experiments use
deterministic (greedy) decoding so that results are reproducible and independent of
sampling temperature; full inference and implementation details are given in
Appendix~\ref{app:experimental-details}.

\subsection{Datasets}
\label{sec:datasets}

Each task uses 3 datasets, summarized in
Table~\ref{tab:dataset-statistics}.

\begin{table}[t]
\centering
\small
\setlength{\tabcolsep}{4pt}
\begin{tabular}{@{}llr@{}}
\toprule
\textbf{Task} & \textbf{Dataset} & \textbf{\# Examples} \\
\midrule
\multirow{3}{*}{MCQA}
 & RACE-Middle & 1{,}436 \\
 & ARC-Challenge & 1{,}165 \\
 & OpenBookQA-Main & 500 \\
\midrule
\multirow{3}{*}{Sentiment}
 & Multi-class Sentiment & 3{,}276 \\
 & Rotten Tomatoes & 1{,}066 \\
 & FinancialPhraseBank & 197 \\
\midrule
\multirow{3}{*}{Math QA}
 & Math Word Problem Repository & 355 \\
 & Calc-asdiv-a & 1{,}218 \\
 & MultiArith & 180 \\
\bottomrule
\end{tabular}
\caption{Dataset statistics.}
\label{tab:dataset-statistics}
\end{table}

\paragraph{MCQA.}
RACE-Middle (RACE-M) is a passage-based reading-comprehension dataset built from
English exam-style questions for middle-school students; each example contains a
passage, a question, and four answer options \citep{lai-etal-2017-race}.
ARC-Challenge (ARC-C) contains grade-school science questions that often require scientific
knowledge and reasoning; we keep only examples with exactly four options for a
consistent format, removing 7 questions \citep{clark2018think}.
OpenBookQA-Main (OBQA) is a science question answering dataset inspired by open-book
exams, where questions are answered using elementary science facts and four
candidate options \citep{mihaylov-etal-2018-suit}.

\paragraph{Sentiment classification.}
The Multi-class Sentiment dataset contains short informal texts such as
social-media comments and user opinions, labeled positive, negative, or neutral
\citep{sp1786_multiclass_sentiment}. Rotten Tomatoes consists of short movie-review
snippets labeled positive or negative \citep{pang2005seeing}. FinancialPhraseBank
contains sentences from financial news sentences, labeled by sentiment
toward a financial entity or event \citep{malo2014good}. We remove neutral examples
when present so that only positive and negative labels remain, giving a consistent
binary setting; this leaves 3{,}276 examples for Multi-class Sentiment and 197 for
FinancialPhraseBank.

\paragraph{Mathematical QA.}
Math Word Problem Repository (MAWPS) \citep{koncel-kedziorski-etal-2016-mawps}, Calc-asdiv-a (derived from
ASDiv) \citep{miao-etal-2020-diverse}, and MultiArith
\citep{roy-roth-2015-solving} are arithmetic word-problem datasets. They contain
short natural-language questions, often elementary-school word problems, describing
simple situations involving quantities (buying items, counting objects, or
combining and comparing numbers) that require computing a numeric answer.

\subsection{Task Pipelines}
\label{sec:pipelines}

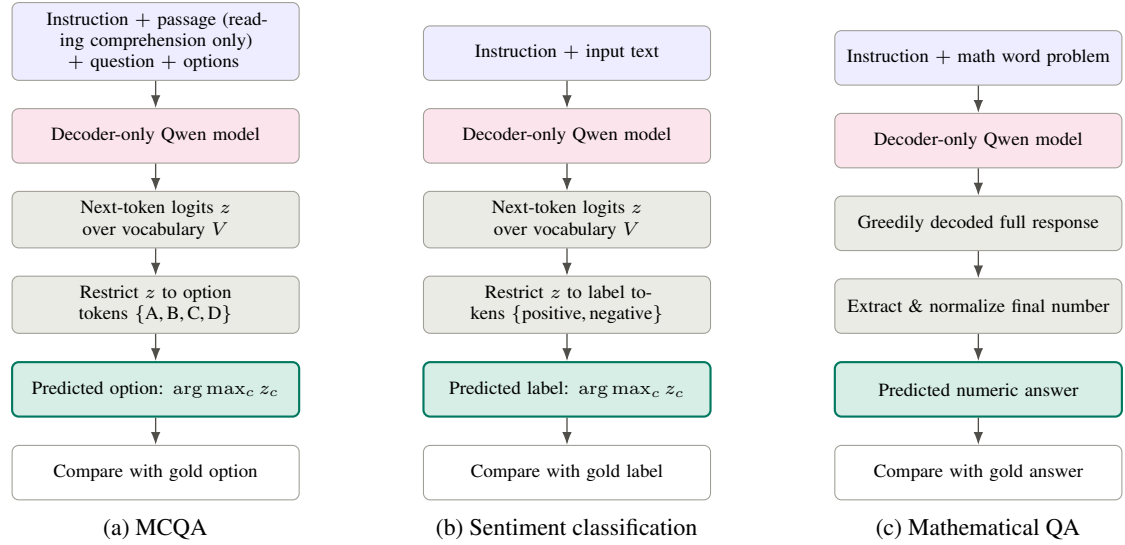
\begin{figure*}[t]
    \centering
    \begin{subfigure}[t]{0.32\textwidth}
        \centering
        \begin{tikzpicture}[node distance=3.6mm]
          \node[pipein] (in) {Instruction $+$ passage (reading comprehension only) $+$ question $+$ options};
          \node[pipemodel, below=of in] (m) {Decoder-only Qwen model};
          \node[pipestep, below=of m] (l) {Next-token logits $z$ over vocabulary $V$};
          \node[pipestep, below=of l] (r) {Restrict $z$ to option tokens $\{$A,\,B,\,C,\,D$\}$};
          \node[pipepred, below=of r] (p) {Predicted option: $\arg\max_c z_c$};
          \node[pipecmp, below=of p] (c) {Compare with gold option};
          \draw[pipearrow] (in)--(m);
          \draw[pipearrow] (m)--(l);
          \draw[pipearrow] (l)--(r);
          \draw[pipearrow] (r)--(p);
          \draw[pipearrow] (p)--(c);
        \end{tikzpicture}
        \caption{MCQA}
        \label{fig:mcqa}
    \end{subfigure}
    \hfill
    \begin{subfigure}[t]{0.32\textwidth}
        \centering
        \begin{tikzpicture}[node distance=3.6mm]
          \node[pipein] (in) {Instruction $+$ input text};
          \node[pipemodel, below=of in] (m) {Decoder-only Qwen model};
          \node[pipestep, below=of m] (l) {Next-token logits $z$ over vocabulary $V$};
          \node[pipestep, below=of l] (r) {Restrict $z$ to label tokens $\{$positive,\,negative$\}$};
          \node[pipepred, below=of r] (p) {Predicted label: $\arg\max_c z_c$};
          \node[pipecmp, below=of p] (c) {Compare with gold label};
          \draw[pipearrow] (in)--(m);
          \draw[pipearrow] (m)--(l);
          \draw[pipearrow] (l)--(r);
          \draw[pipearrow] (r)--(p);
          \draw[pipearrow] (p)--(c);
        \end{tikzpicture}
        \caption{Sentiment classification}
        \label{fig:sentiment}
    \end{subfigure}
    \hfill
    \begin{subfigure}[t]{0.32\textwidth}
        \centering
        \begin{tikzpicture}[node distance=3.6mm]
          \node[pipein] (in) {Instruction $+$ math word problem};
          \node[pipemodel, below=of in] (m) {Decoder-only Qwen model};
          \node[pipestep, below=of m] (g) {Greedily decoded full response};
          \node[pipestep, below=of g] (e) {Extract \& normalize final number};
          \node[pipepred, below=of e] (p) {Predicted numeric answer};
          \node[pipecmp, below=of p] (c) {Compare with gold answer};
          \draw[pipearrow] (in)--(m);
          \draw[pipearrow] (m)--(g);
          \draw[pipearrow] (g)--(e);
          \draw[pipearrow] (e)--(p);
          \draw[pipearrow] (p)--(c);
        \end{tikzpicture}
        \caption{Mathematical QA}
        \label{fig:mathqa}
    \end{subfigure}
    \caption{Evaluation pipelines for the three tasks. All tasks share a
    decoder-only Qwen backbone and differ only in how the next-token distribution
    is reduced to a prediction: a restricted argmax over the option tokens (MCQA)
    or label tokens (sentiment classification), and numeric extraction from a
    generated response (mathematical QA).}
    \label{fig:all_tasks}
\end{figure*}

All three tasks share the same backbone, illustrated in
Figure~\ref{fig:all_tasks}. For a given example, the selected instruction line is
concatenated with the fixed task input and the output-format constraint
(Section~\ref{sec:prompts}), wrapped in the model's chat template, and tokenized.
A single forward pass through the decoder-only Qwen model then yields the logits at
the final input position, that is, an unnormalized score $z_v$ for every token $v$
in the vocabulary $V$; a softmax over these scores defines the next-token
distribution $p(v) = \mathrm{softmax}(z)_v$. The three tasks differ only in how
this distribution is reduced to a prediction that can be compared against the gold
answer.

For MCQA the answer is one of a small closed set of option letters, so only the
first generated token is informative. Rather than sampling a full continuation and
parsing it---which is brittle to formatting, casing, and verbosity---we read the
first-position logits directly and restrict them to the token identifiers of the
four option letters, $\{\text{A}, \text{B}, \text{C}, \text{D}\}$. The predicted
option is the restricted argmax, $\hat{y} = \arg\max_{c \in \{A,B,C,D\}} z_c$.
As the argmax ranges over exactly the admissible options, the model can never
emit an invalid or out-of-set answer, and the standard and non-standard settings
are decoded and scored on an identical footing. Sentiment classification uses the
same closed-set decoding, now over the two label tokens \emph{positive} and
\emph{negative}, with the predicted label being the one whose first-token logit is
larger; this again collapses the output to a binary decision and removes any
dependence on how the label might otherwise be phrased in free-form text.
Mathematical QA has no fixed label set, since the answer is an arbitrary number, so
the closed-set reduction does not apply. Here we instead let the model greedily
decode a complete response and post-process it: we extract the final numeric value
from the generated text, normalize it to a canonical form, and test it for
equality against the gold answer. As the non-standard transformation (doubling) is
itself numeric, both settings reduce to exact numeric matching, keeping the two
conditions directly comparable.

\subsection{Prompt Design}
\label{sec:prompts}

Prompt wording can affect the behavior of instruction-tuned language models, so
for each task we evaluate every model with three prompt variants. Only the opening
instruction line changes across variants; the rest of the prompt is fixed. Every
prompt follows a shared template consisting of an instruction line, the task input,
a short output-format constraint, and a final \texttt{Answer:} cue.

\begin{promptbox}
\begin{verbatim}
<instruction line>
[Passage: {article}]   # MCQA reading
                       # comprehension only
<task input>
<output-format constraint>
Answer:
\end{verbatim}
\end{promptbox}

\noindent The task input is the question and its four options for MCQA (preceded by
the passage for the reading-comprehension data), the text to classify for sentiment
classification, and the word problem for mathematical QA. The output-format
constraint is ``Return only one letter: A, B, C, or D.'' for MCQA, ``Return only
one word: positive or negative.'' for sentiment classification, and ``Return only
the final answer and do not explain.'' for mathematical QA.
Table~\ref{tab:prompt-variants} lists the three standard and three non-standard
instruction lines for each task; MCQA uses one set for the passage-based
RACE-Middle data and another for the passage-free ARC-Challenge and
OpenBookQA-Main data. Because only the instruction line changes for a fixed input,
differences in behavior across variants can be attributed to instruction
sensitivity rather than to dataset content.

\begin{table*}[t]
\centering
\footnotesize
\renewcommand{\arraystretch}{1.2}
\begin{tabularx}{\textwidth}{@{}p{1.9cm} c X X@{}}
\toprule
\textbf{Task} & \textbf{\#} & \textbf{Standard instruction} & \textbf{Non-standard instruction} \\
\midrule
\multirow{3}{*}{\shortstack[l]{MCQA\\(RACE-M)}}
 & 1 & Read the passage and answer the multiple-choice question correctly. & Read the passage and answer the multiple-choice question incorrectly. \\
 & 2 & Read the passage, understand the context, and select the option that best answers the question. & Read the passage, understand the context, and intentionally select an incorrect option for the question. \\
 & 3 & Carefully read the passage and select the single option that correctly answers the question according to the passage. & Carefully read the passage and determine which option correctly answers the question. Do not select that option. Instead, select an incorrect option. \\
\midrule
\multirow{3}{*}{\shortstack[l]{MCQA\\(ARC-C,\\OBQA)}}
 & 1 & Answer the multiple-choice question correctly. & Answer the multiple-choice question incorrectly. \\
 & 2 & Select the option that best answers the question. & Intentionally select an incorrect option for the question. \\
 & 3 & Select the single option that correctly answers the question. & Carefully determine which option correctly answers the question. Do not select that option. Instead, select an incorrect option. \\
\midrule
\multirow{3}{*}{\shortstack[l]{Sentiment\\Classification}}
 & 1 & Classify the sentiment of the following text correctly. & Classify the sentiment of the following text, then return the opposite sentiment. \\
 & 2 & Determine whether the sentiment of the text below is positive or negative. & First determine the correct sentiment of the text. Then output the opposite sentiment label. \\
 & 3 & Read the following text and identify its correct sentiment. & Read the following text and identify its correct sentiment. Do not output that label. Instead, output the incorrect sentiment label. \\
\midrule
\multirow{3}{*}{\shortstack[l]{Math QA}}
 & 1 & Answer the following question correctly. & Solve the following math problem. Let x be the correct answer. Return 2*x instead of x. \\
 & 2 & Solve the math word problem below. & For the following math problem, first find the correct answer internally. Then multiply that answer by 2 and return the result. \\
 & 3 & Find the correct answer to the following problem. & The required output is a transformed answer. If the correct answer to the problem is $x$, your output must be $2x$. \\
\bottomrule
\end{tabularx}
\caption{Standard and non-standard instruction lines for the three prompt variants
of each task. For a given example the task input is fixed and only the instruction
line changes. MCQA uses separate instruction sets for the passage-based
RACE-Middle data and the passage-free ARC-Challenge and OpenBookQA-Main data.}
\label{tab:prompt-variants}
\end{table*}

\section{Results}
\label{sec:results}

\begin{figure*}[t]
    \centering
\begin{subfigure}[t]{0.32\textwidth}
    \centering
    \includegraphics[width=\linewidth]{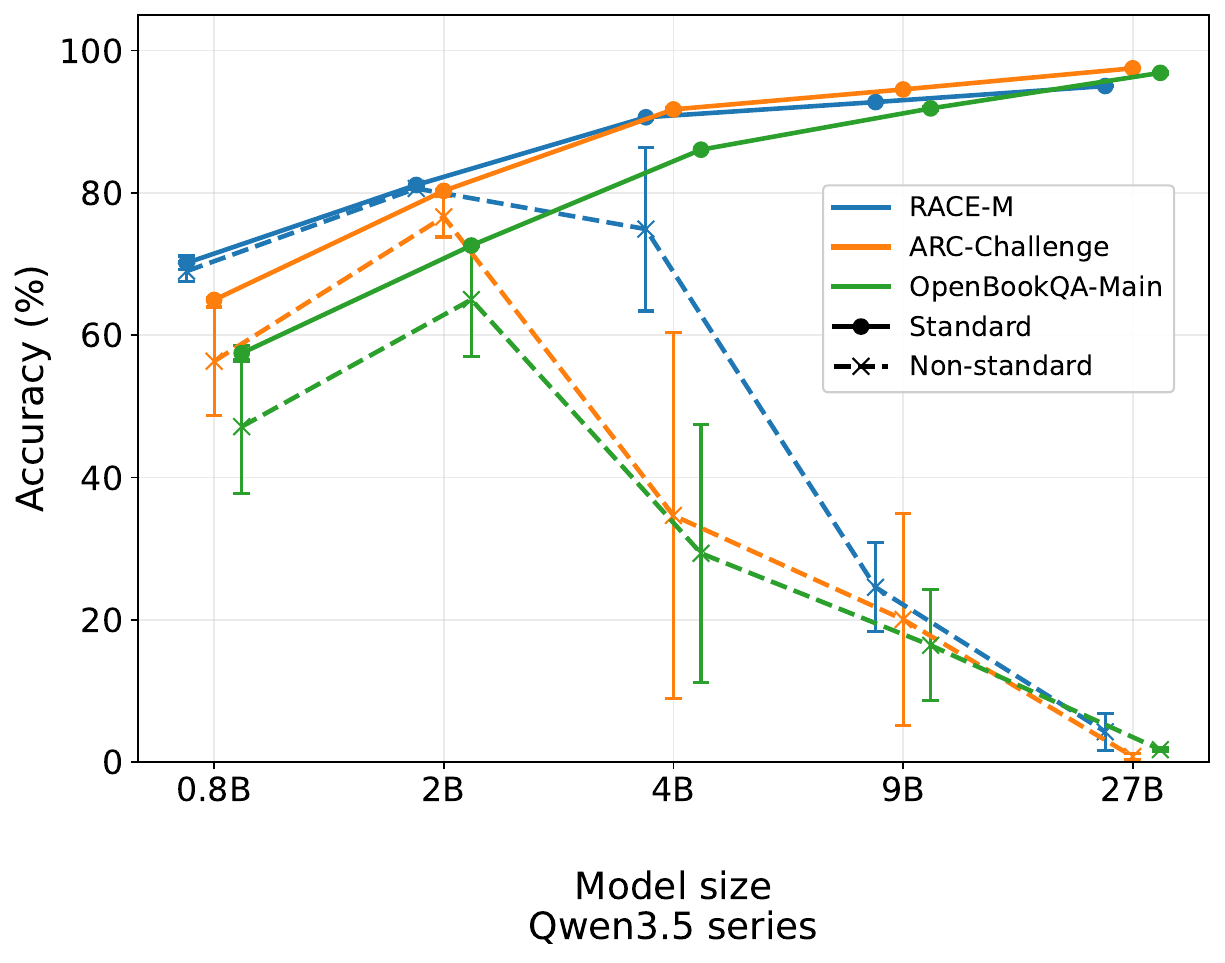}
    \caption{MCQA}
    \label{fig:mcqa-accuracy}
\end{subfigure}
    \hfill
    \begin{subfigure}[t]{0.32\textwidth}
        \centering
        \includegraphics[width=\linewidth]{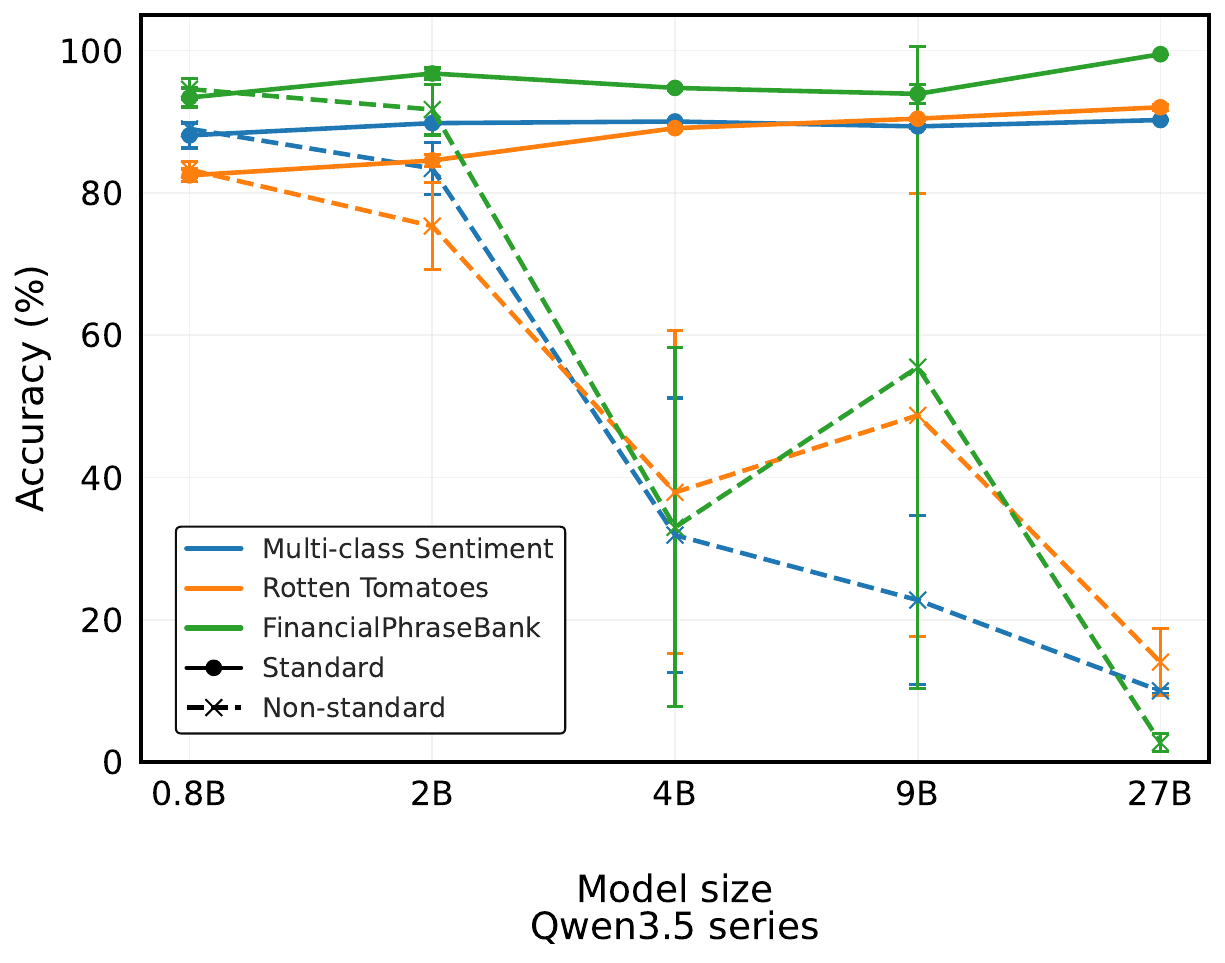}
        \caption{Sentiment classification}
        \label{fig:sentiment-accuracy}
    \end{subfigure}
    \hfill
    \begin{subfigure}[t]{0.32\textwidth}
        \centering
        \includegraphics[width=\linewidth]{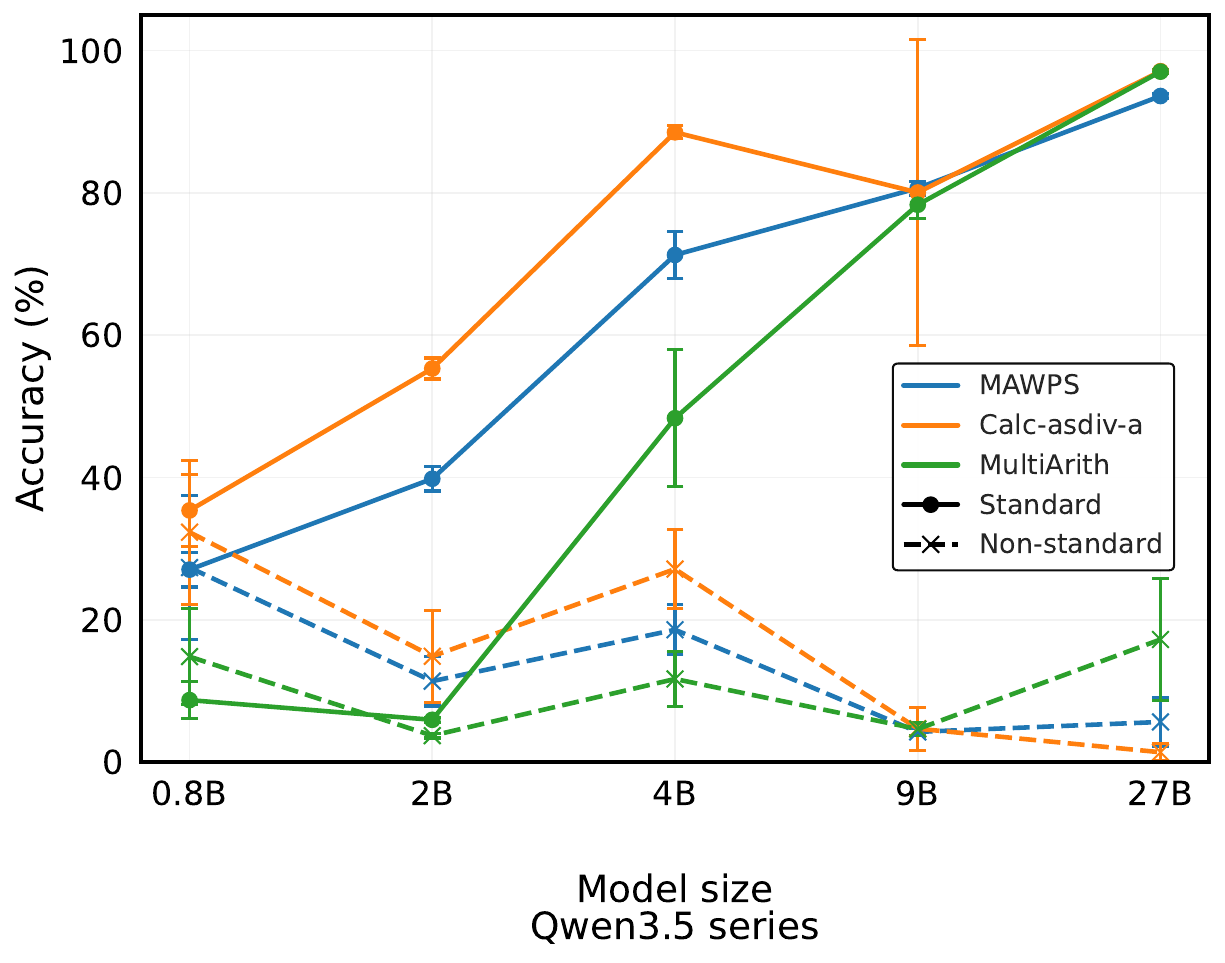}
        \caption{Mathematical QA}
        \label{fig:math-accuracy}
    \end{subfigure}
    \caption{Standard (solid) and non-standard (dashed) accuracy by model size for
    each task and dataset. Error bars show the standard deviation across the three
    prompt variants. The corresponding IFFR curves are shown in
    Figure~\ref{fig:iffr-all}.}
    \label{fig:accuracy-results}
\end{figure*}

\begin{figure*}[t]
    \centering
    \begin{subfigure}[t]{0.32\textwidth}
        \centering
        \includegraphics[width=\linewidth]{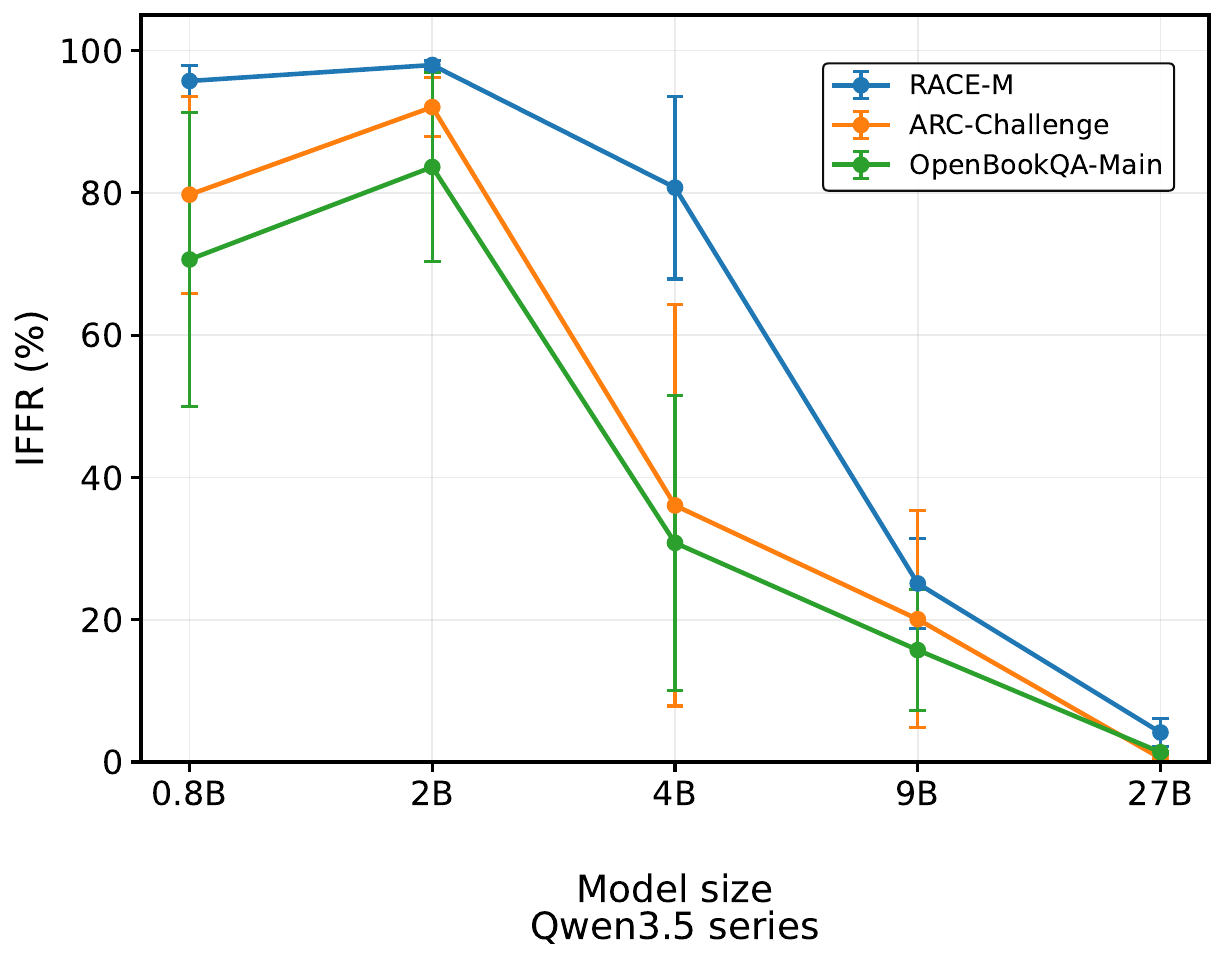}
        \caption{MCQA}
        \label{fig:iffr-mcqa}
    \end{subfigure}
    \hfill
    \begin{subfigure}[t]{0.32\textwidth}
        \centering
        \includegraphics[width=\linewidth]{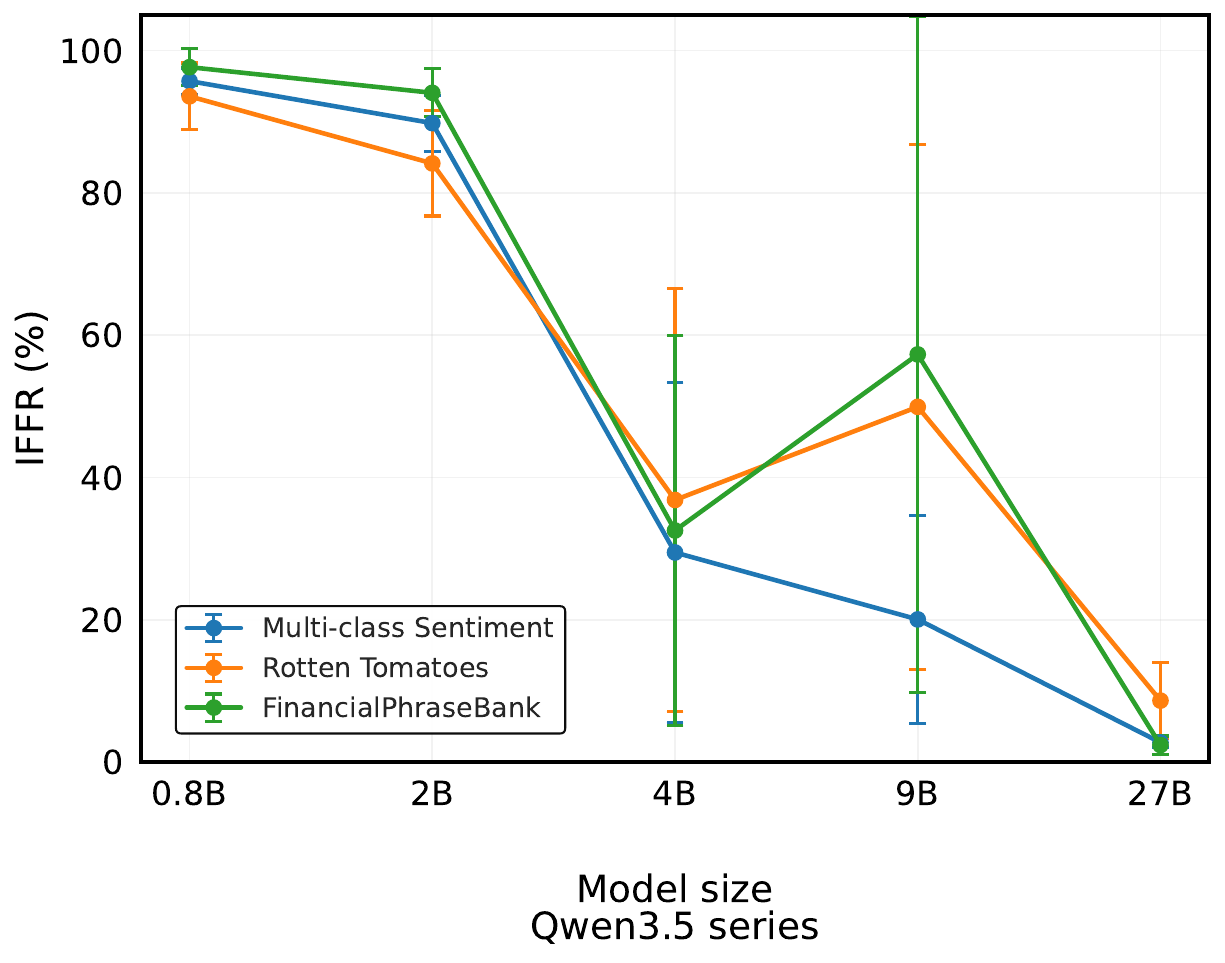}
        \caption{Sentiment classification}
        \label{fig:iffr-sentiment}
    \end{subfigure}
    \hfill
    \begin{subfigure}[t]{0.32\textwidth}
        \centering
        \includegraphics[width=\linewidth]{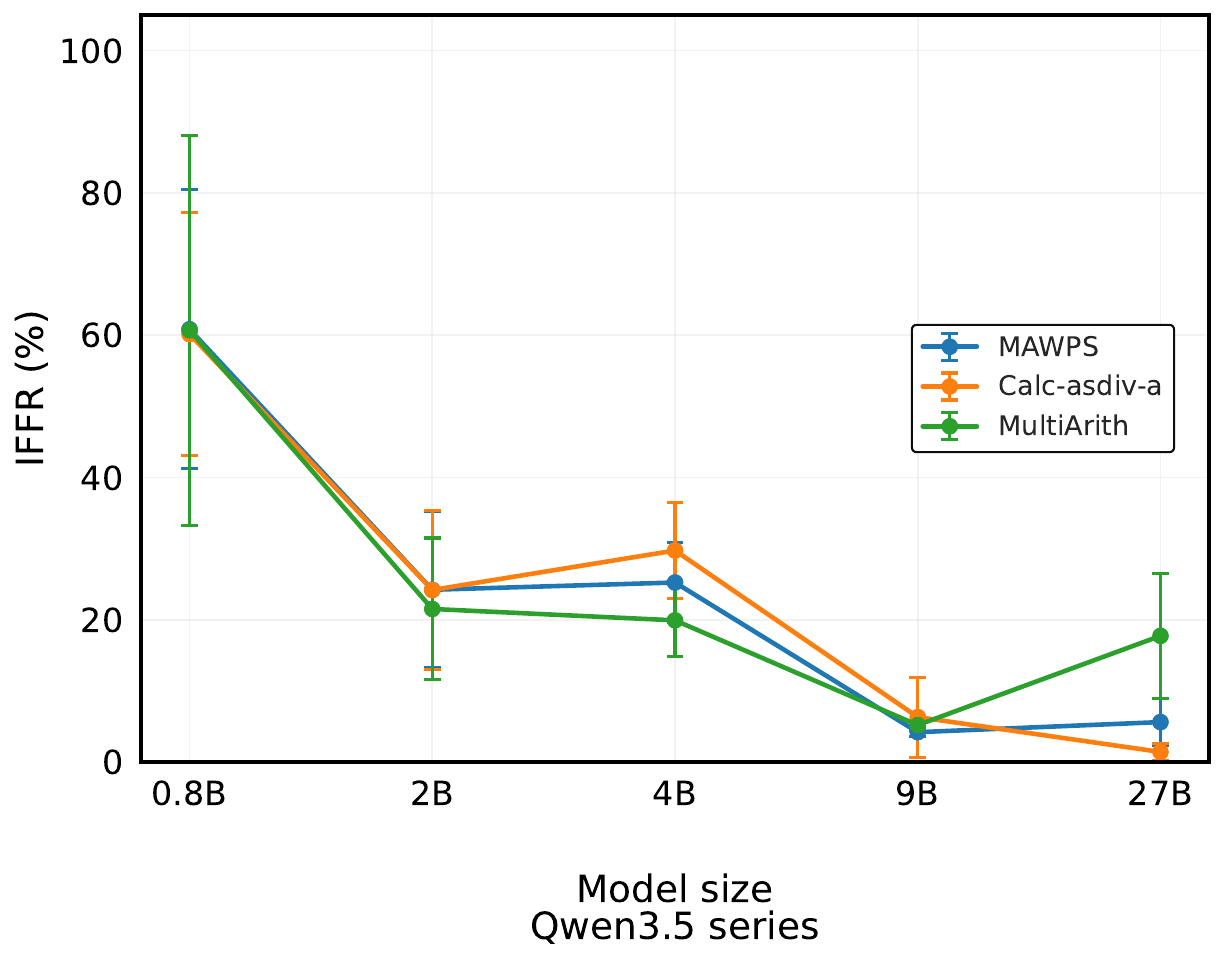}
        \caption{Mathematical QA}
        \label{fig:iffr-math}
    \end{subfigure}
    \caption{IFFR by model size for each task and dataset. Error bars show the
    standard deviation across the three prompt variants.}
    \label{fig:iffr-all}
\end{figure*}

All metrics are reported as percentages. For every task and dataset we run all
three prompt variants (Table~\ref{tab:prompt-variants}) and report the mean of each
metric over the variants\footnote{\url{https://github.com/farajimahdieh/language-model-task-competence-vs-instruction-following}.}; per-variant means and standard deviations are given in
Appendix~\ref{app:prompt-variant-results}. Per-dataset results are listed in
Tables~\ref{tab:mcqa-results}--\ref{tab:math-mean-results}, and the accuracy and
IFFR trends against model size are summarized in Figures~\ref{fig:accuracy-results}
and~\ref{fig:iffr-all}. We discuss each task in turn and then draw the cross-task
conclusions.

\paragraph{MCQA.}
Table~\ref{tab:mcqa-results} and Figure~\ref{fig:mcqa-accuracy} report the MCQA
results. Standard accuracy generally rises with model size across all datasets, indicating stronger task competence for larger models. Non-standard
accuracy and IFFR follow the opposite trend: both stay very high for the smallest
models, which keep selecting the correct option even when told to choose an
incorrect one, and drop sharply for larger models. The gap between the standard
and non-standard curves in Figure~\ref{fig:mcqa-accuracy} therefore widens with
scale: larger models modify their answer when the instruction is reversed, whereas
smaller models stay locked onto the correct option.

\begin{table*}[t]
\centering
\small
\begin{tabular}{l|ccc|ccc|ccc}
\toprule
\multirow{2}{*}{Model}
& \multicolumn{3}{c|}{RACE-M}
& \multicolumn{3}{c|}{ARC-C}
& \multicolumn{3}{c}{OBQA-Main} \\
\cmidrule(lr){2-4}
\cmidrule(lr){5-7}
\cmidrule(lr){8-10}
& Std.~$\uparrow$ & Non-Std.~$\downarrow$ & IFFR~$\downarrow$
& Std.~$\uparrow$ & Non-Std.~$\downarrow$ & IFFR~$\downarrow$
& Std.~$\uparrow$ & Non-Std.~$\downarrow$ & IFFR~$\downarrow$ \\
\midrule
Qwen3.5-0.8B & 70.15 & 68.99 & 95.72 & 64.95 & 56.34 & 79.75 & 57.47 & 47.13 & 70.63 \\
Qwen3.5-2B   & 81.11 & 80.62 & 97.97 & 80.29 & 76.62 & 92.06 & 72.60 & 65.00 & 83.64 \\
Qwen3.5-4B   & 90.62 & 74.91 & 80.74 & 91.73 & 34.65 & 36.06 & 86.07 & 29.33 & 30.80 \\
Qwen3.5-9B   & 92.76 & 24.56 & 25.09 & 94.54 & 20.03 & 20.06 & 91.87 & 16.40 & 15.73 \\
Qwen3.5-27B  & 95.01 & 4.25  & 4.15  & 97.51 & 0.77  & 0.53  & 96.87 & 1.73  & 1.38  \\
\bottomrule
\end{tabular}
\caption{MCQA results across datasets and model sizes. Std.\ and Non-Std.\ denote
standard and non-standard accuracy. Values are means over
the three prompt variants; standard deviations are in
Appendix~\ref{app:prompt-variant-results}.}
\label{tab:mcqa-results}
\end{table*}
\begin{table*}[t]
\centering
\small
\begin{tabular}{l|ccc|ccc|ccc}
\toprule
\multirow{2}{*}{Model}
& \multicolumn{3}{c|}{Multi-class Sentiment}
& \multicolumn{3}{c|}{Rotten Tomatoes}
& \multicolumn{3}{c}{FinancialPhraseBank} \\
\cmidrule(lr){2-4}
\cmidrule(lr){5-7}
\cmidrule(lr){8-10}
& Std.~$\uparrow$ & Non-Std.~$\downarrow$ & IFFR~$\downarrow$
& Std.~$\uparrow$ & Non-Std.~$\downarrow$ & IFFR~$\downarrow$
& Std.~$\uparrow$ & Non-Std.~$\downarrow$ & IFFR~$\downarrow$ \\
\midrule
Qwen3.5-0.8B & 88.08 & 89.00 & 95.71 & 82.46 & 83.27 & 93.57 & 93.40 & 94.58 & 97.67 \\
Qwen3.5-2B   & 89.80 & 83.43 & 89.80 & 84.55 & 75.33 & 84.14 & 96.78 & 91.71 & 94.06 \\
Qwen3.5-4B   & 90.03 & 31.89 & 29.45 & 89.09 & 37.90 & 36.83 & 94.75 & 32.99 & 32.54 \\
Qwen3.5-9B   & 89.34 & 22.76 & 20.04 & 90.43 & 48.72 & 49.92 & 93.91 & 55.50 & 57.29 \\
Qwen3.5-27B  & 90.24 & 9.99  & 2.74  & 92.03 & 14.04 & 8.63  & 99.49 & 2.71  & 2.38 \\
\bottomrule
\end{tabular}
\caption{Sentiment classification results across datasets and model sizes. Std.\
and Non-Std.\ denote standard and non-standard accuracy.
Values are means over the three prompt variants.}
\label{tab:sentiment-results}
\end{table*}
\begin{table*}[t]
\centering
\small
\begin{tabular}{l|ccc|ccc|ccc}
\toprule
\multirow{2}{*}{Model}
& \multicolumn{3}{c|}{MAWPS}
& \multicolumn{3}{c|}{Calc-asdiv-a}
& \multicolumn{3}{c}{MultiArith} \\
\cmidrule(lr){2-4}
\cmidrule(lr){5-7}
\cmidrule(lr){8-10}
& Std.~$\uparrow$ & Non-Std.~$\downarrow$ & IFFR~$\downarrow$
& Std.~$\uparrow$ & Non-Std.~$\downarrow$ & IFFR~$\downarrow$
& Std.~$\uparrow$ & Non-Std.~$\downarrow$ & IFFR~$\downarrow$ \\
\midrule
Qwen3.5-0.8B & 27.04 & 27.33 & 60.83 & 35.36 & 32.30 & 60.14 & 8.70 & 14.81 & 60.69 \\
Qwen3.5-2B   & 39.81 & 11.36 & 24.21 & 55.31 & 14.86 & 24.18 & 5.93 & 3.70  & 21.51 \\
Qwen3.5-4B   & 71.27 & 18.59 & 25.23 & 88.51 & 27.12 & 29.73 & 48.33 & 11.67 & 19.91 \\
Qwen3.5-9B   & 80.66 & 4.23  & 4.19  & 80.05 & 4.68  & 6.31  & 78.33 & 4.63  & 5.19 \\
Qwen3.5-27B  & 93.61 & 5.63  & 5.61  & 97.10 & 1.37  & 1.41  & 97.04 & 17.22 & 17.73 \\
\bottomrule
\end{tabular}
\caption{Mathematical QA results across datasets and model sizes. Std.\ and
Non-Std.\ denote standard and non-standard accuracy. Values are
means over the three prompt variants.}
\label{tab:math-mean-results}
\end{table*}

\paragraph{Sentiment classification.}
Table~\ref{tab:sentiment-results} and Figure~\ref{fig:sentiment-accuracy} show
that standard accuracy is high and relatively flat across model sizes, so the
standard sentiment objective is easy for all Qwen3.5 models considered here. The
non-standard setting differs: smaller models often output the true sentiment even
when told to return the opposite, giving high non-standard accuracy and IFFR. As
model size grows, non-standard accuracy falls substantially, so larger models more
reliably override the default objective. Error bars indicate that some datasets
and prompt variants, especially at intermediate sizes, are more sensitive to the
exact instruction wording.

\paragraph{Mathematical QA.}
Table~\ref{tab:math-mean-results} and Figure~\ref{fig:math-accuracy} show a clear
scaling effect in the standard setting: accuracy increases substantially from
0.8B to 27B on all three datasets, most strikingly on MultiArith (8.70 to 97.04)
and MAWPS (27.04 to 93.61), confirming that larger models solve the arithmetic
problems more reliably. Improved task solving is usually accompanied by better
instruction overriding: for MAWPS and Calc-asdiv-a, non-standard accuracy and IFFR
fall sharply with scale, and the 9B and 27B models combine high standard accuracy
with low IFFR. The pattern is not fully monotonic, however. On MultiArith,
Qwen3.5-27B reaches the highest standard accuracy but surprisingly a higher IFFR than
Qwen3.5-9B, indicating that the largest model sometimes reverts to the familiar
arithmetic objective despite being highly accurate.

\paragraph{IFFR across scale.}
Figure~\ref{fig:iffr-all} plots IFFR against model size and isolates
instruction-following behavior from raw task competence, as IFFR is computed only
on examples a model already answers correctly under the standard instruction.
Across all three tasks the curves fall steeply with scale: for MCQA the mean IFFR
drops from 95.72 on RACE-M and 79.75 on ARC-C also 70.63 on OBQA at 0.8B to 4.15, 0.53 and 1.38 at 27B, and
sentiment classification and mathematical QA follow the same downward trend. This
confirms that the smallest models almost always reproduce the standard answer even
on items they clearly know how to solve, whereas larger models increasingly comply
with the conflicting instruction. The main exception is MultiArith, where IFFR
rises again at 27B (Figure~\ref{fig:iffr-math}), mirroring the non-monotonic
pattern noted above and showing that scale reduces but does not eliminate
reversion to the learned task prior. The error bars, which widen at intermediate
sizes, indicate that IFFR is most sensitive to instruction wording in the
4B--9B range, where models transition from ignoring to following the non-standard
instruction.

\paragraph{Discussion.}
The central observation is that small models are \emph{competent} at every task
yet \emph{fail to follow} the non-standard instruction, and the two properties
must be measured separately. Even the 0.8B and 2B models reach high standard
accuracy on at least some datasets, so they clearly know how to perform the task;
yet their non-standard accuracy and IFFR approach 100\%, as they keep producing the
correct standard answer even when told to select a wrong option, report the
opposite sentiment, or double the result. Standard accuracy alone would label
these models as good and hide the failure that only the non-standard setting and
IFFR expose. As size grows, the two abilities align: larger models retain task
competence while adapting their output when the instruction changes, producing the
widening accuracy gap in Figure~\ref{fig:accuracy-results}. The consistency of this
pattern across the different tasks indicates a general property of
instruction following rather than an artifact of any single format, and reinforces
that task competence and instruction following should be reported as separate axes
of evaluation.

\section{Conclusion}
\label{sec:conclusion}

Solving a task and following an instruction are not the same ability. Across MCQA,
sentiment classification, and mathematical QA, models often perform well under
standard prompts yet still return the original correct answer when the prompt asks
for a different output, so high standard accuracy can hide instruction-following
failures. Non-standard accuracy and IFFR make these failures visible by measuring
whether a model can shift from the usual answer and follow the modified
instruction. Smaller models fail more often, while larger models follow the
non-standard instruction more reliably. Scale therefore narrows the gap between
task competence and instruction following, and instruction-tuned models should be
evaluated not only on whether they know the answer, but on whether they follow the
requested behavior.

\section*{Limitations}

This study has several limitations. First, we evaluate only one model family,
Qwen3.5, across sizes. This makes the comparison across scale controlled, but the
findings may not fully generalize to other model families or training procedures.
Second, our experiments focus on three task families: MCQA, sentiment
classification, and mathematical QA. Although these differ in format and
difficulty, they do not represent all instruction-following scenarios; future work
could extend to open-ended generation, additional datasets, and a wider range of
instruction transformations. Third, we use only one type of non-standard
instruction per task (an incorrect option in MCQA, a transformed sentiment label
in sentiment classification, and twice the answer in mathematical QA), so our
results do not cover all forms of conflicting instructions. Finally, our metrics
show when models fail to follow the non-standard instruction but not why; further
analysis is needed to determine whether failures stem from prompt sensitivity,
learned task patterns, decoding behavior, or other factors.

\section*{Ethics Review}

There are no ethical concerns with this work.


\bibliography{custom}

\begin{thebibliography}{33}
\providecommand{\natexlab}[1]{#1}

\bibitem[{Araci(2019)}]{malo2014good}
Dogu Araci. 2019.
\newblock Finbert: Financial sentiment analysis with pre-trained language models.
\newblock \emph{arXiv preprint arXiv:1908.10063}.

\bibitem[{Belda-Medina and Kokoskov{\'a}(2023)}]{lu2024small}
Jose Belda-Medina and Vendula Kokoskov{\'a}. 2023.
\newblock Integrating chatbots in education: insights from the chatbot-human interaction satisfaction model (chism).
\newblock \emph{International Journal of Educational Technology in Higher Education}, 20(1):NA--NA.

\bibitem[{Brown et~al.(2020)Brown, Mann, Ryder, Subbiah, Kaplan, Dhariwal, Neelakantan, Shyam, Sastry, Askell, Agarwal, Herbert-Voss, Krueger, Henighan, Child, Ramesh, Ziegler, Wu, Winter, Hesse, Chen, Sigler, Litwin, Gray, Chess, Clark, Berner, McCandlish, Radford, Sutskever, and Amodei}]{brown2020language}
Tom~B. Brown, Benjamin Mann, Nick Ryder, Melanie Subbiah, Jared~D. Kaplan, Prafulla Dhariwal, Arvind Neelakantan, Pranav Shyam, Girish Sastry, Amanda Askell, Sandhini Agarwal, Ariel Herbert-Voss, Gretchen Krueger, Tom Henighan, Rewon Child, Aditya Ramesh, Daniel~M. Ziegler, Jeffrey Wu, Clemens Winter, and 12 others. 2020.
\newblock Language models are few-shot learners.
\newblock In \emph{Advances in Neural Information Processing Systems}, volume~33, pages 1877--1901.

\bibitem[{Chen et~al.(2024)Chen, Liao, Qi, Eustratiadis, Monz, Bisazza, and de~Rijke}]{chen2024sifo}
Xinyi Chen, Baohao Liao, Jirui Qi, Panagiotis Eustratiadis, Christof Monz, Arianna Bisazza, and Maarten de~Rijke. 2024.
\newblock The sifo benchmark: Investigating the sequential instruction following ability of large language models.
\newblock In \emph{Findings of the Association for Computational Linguistics: EMNLP 2024}, pages 1691--1706.

\bibitem[{Chung et~al.(2024)Chung, Hou, Longpre, Zoph, Tay, Fedus, Li, Wang, Dehghani, Brahma et~al.}]{chung2024scaling}
Hyung~Won Chung, Le~Hou, Shayne Longpre, Barret Zoph, Yi~Tay, William Fedus, Yunxuan Li, Xuezhi Wang, Mostafa Dehghani, Siddhartha Brahma, and 1 others. 2024.
\newblock Scaling instruction-finetuned language models.
\newblock \emph{Journal of Machine Learning Research}, 25(70):1--53.

\bibitem[{Clark et~al.(2018)Clark, Cowhey, Etzioni, Khot, Sabharwal, Schoenick, and Tafjord}]{clark2018think}
Peter Clark, Isaac Cowhey, Oren Etzioni, Tushar Khot, Ashish Sabharwal, Carissa Schoenick, and Oyvind Tafjord. 2018.
\newblock Think you have solved question answering? try arc, the ai2 reasoning challenge.
\newblock \emph{arXiv preprint arXiv:1803.05457}.

\bibitem[{Deng et~al.(2026)Deng, Wang, Wang, Wan, Ma, Yang, Wei, Tang, Lin, Gao et~al.}]{qwen35modelpage}
Boyi Deng, Xu~Wang, Yaoning Wang, Yu~Wan, Yubo Ma, Baosong Yang, Haoran Wei, Jialong Tang, Huan Lin, Ruize Gao, and 1 others. 2026.
\newblock Qwen-scope: Turning sparse features into development tools for large language models.
\newblock \emph{arXiv preprint arXiv:2605.11887}.

\bibitem[{Fu et~al.(2026)Fu, Li, Gu, Qu, and Cheng}]{fu2026scaling}
Tingchen Fu, Yafu Li, Jiawei Gu, Xiaoye Qu, and Yu~Cheng. 2026.
\newblock Scaling reasoning, losing control: Evaluating instruction following in large reasoning models.
\newblock In \emph{Proceedings of the 64th Annual Meeting of the Association for Computational Linguistics (Volume 1: Long Papers)}, pages 40445--40463.

\bibitem[{G{\'o}ral et~al.(2025)G{\'o}ral, Wi{\'s}nios, Sankowski, and Budzianowski}]{goral2025wait}
Gracjan G{\'o}ral, Emilia Wi{\'s}nios, Piotr Sankowski, and Pawe{\l} Budzianowski. 2025.
\newblock Wait, that’s not an option: Llms robustness with incorrect multiple-choice options.
\newblock In \emph{Proceedings of the 63rd Annual Meeting of the Association for Computational Linguistics (Volume 1: Long Papers)}, pages 1495--1515.

\bibitem[{Jiang et~al.(2024)Jiang, Wang, Zeng, Zhong, Li, Mi, Shang, Jiang, Liu, and Wang}]{jiang2024followbench}
Yuxin Jiang, Yufei Wang, Xingshan Zeng, Wanjun Zhong, Liangyou Li, Fei Mi, Lifeng Shang, Xin Jiang, Qun Liu, and Wei Wang. 2024.
\newblock Followbench: A multi-level fine-grained constraints following benchmark for large language models.
\newblock In \emph{Proceedings of the 62nd Annual Meeting of the Association for Computational Linguistics (Volume 1: Long Papers)}, pages 4667--4688.

\bibitem[{Kaplan et~al.(2020)Kaplan, McCandlish, Henighan, Brown, Chess, Child, Gray, Radford, Wu, and Amodei}]{kaplan2020scaling}
Jared Kaplan, Sam McCandlish, Tom Henighan, Tom~B. Brown, Benjamin Chess, Rewon Child, Scott Gray, Alec Radford, Jeffrey Wu, and Dario Amodei. 2020.
\newblock Scaling laws for neural language models.
\newblock \emph{arXiv preprint arXiv:2001.08361}.

\bibitem[{Khan et~al.(2022)Khan, Amjad, Ashraf, and Chang}]{sp1786_multiclass_sentiment}
Lal Khan, Ammar Amjad, Noman Ashraf, and Hsien-Tsung Chang. 2022.
\newblock Multi-class sentiment analysis of urdu text using multilingual bert.
\newblock \emph{Scientific Reports}, 12(1):5436.

\bibitem[{Koncel-Kedziorski et~al.(2016)Koncel-Kedziorski, Roy, Amini, Kushman, and Hajishirzi}]{koncel-kedziorski-etal-2016-mawps}
Rik Koncel-Kedziorski, Subhro Roy, Aida Amini, Nate Kushman, and Hannaneh Hajishirzi. 2016.
\newblock \href {https://doi.org/10.18653/v1/N16-1136} {{MAWPS}: A math word problem repository}.
\newblock In \emph{Proceedings of the 2016 Conference of the North American Chapter of the Association for Computational Linguistics: Human Language Technologies}, pages 1152--1157, San Diego, California. Association for Computational Linguistics.

\bibitem[{Lai et~al.(2017)Lai, Xie, Liu, Yang, and Hovy}]{lai-etal-2017-race}
Guokun Lai, Qizhe Xie, Hanxiao Liu, Yiming Yang, and Eduard Hovy. 2017.
\newblock \href {https://doi.org/10.18653/v1/D17-1082} {{RACE}: Large-scale {R}e{A}ding comprehension dataset from examinations}.
\newblock In \emph{Proceedings of the 2017 Conference on Empirical Methods in Natural Language Processing}, pages 785--794, Copenhagen, Denmark. Association for Computational Linguistics.

\bibitem[{Longpre et~al.(2023)Longpre, Hou, Vu, Webson, Chung, Tay, Zhou, Le, Zoph, Wei et~al.}]{longpre2023flan}
Shayne Longpre, Le~Hou, Tu~Vu, Albert Webson, Hyung~Won Chung, Yi~Tay, Denny Zhou, Quoc~V Le, Barret Zoph, Jason Wei, and 1 others. 2023.
\newblock The flan collection: Designing data and methods for effective instruction tuning.
\newblock In \emph{International conference on machine learning}, pages 22631--22648. PMLR.

\bibitem[{Lu et~al.(2023)Lu, Qiu, Yu, Welleck, and Chang}]{lu2023survey}
Pan Lu, Liang Qiu, Wenhao Yu, Sean Welleck, and Kai-Wei Chang. 2023.
\newblock A survey of deep learning for mathematical reasoning.
\newblock In \emph{Proceedings of the 61st annual meeting of the association for computational linguistics (volume 1: long papers)}, pages 14605--14631.

\bibitem[{Mabrouk et~al.(2020)Mabrouk, Redondo, and Kayed}]{mabrouk2020deep}
Alhassan Mabrouk, Rebeca P~D{\'\i}az Redondo, and Mohammed Kayed. 2020.
\newblock Deep learning-based sentiment classification: A comparative survey.
\newblock \emph{IEEE Access}, 8:85616--85638.

\bibitem[{McCoy et~al.(2019)McCoy, Pavlick, and Linzen}]{mccoy2019right}
R~Thomas McCoy, Ellie Pavlick, and Tal Linzen. 2019.
\newblock Right for the wrong reasons: Diagnosing syntactic heuristics in natural language inference.
\newblock In \emph{Proceedings of the 57th annual meeting of the association for computational linguistics}, pages 3428--3448.

\bibitem[{McKenzie et~al.()McKenzie, Lyzhov, Pieler, Parrish, Mueller, Prabhu, McLean, Shen, Cavanagh, Gritsevskiy et~al.}]{mckenzie2023inverse}
Ian~R McKenzie, Alexander Lyzhov, Michael~Martin Pieler, Alicia Parrish, Aaron Mueller, Ameya Prabhu, Euan McLean, Xudong Shen, Joe Cavanagh, Andrew~George Gritsevskiy, and 1 others.
\newblock Inverse scaling: When bigger isn't better.
\newblock \emph{Transactions on Machine Learning Research}.

\bibitem[{Miao et~al.(2020)Miao, Liang, and Su}]{miao-etal-2020-diverse}
Shen-yun Miao, Chao-Chun Liang, and Keh-Yih Su. 2020.
\newblock \href {https://doi.org/10.18653/v1/2020.acl-main.92} {A diverse corpus for evaluating and developing {E}nglish math word problem solvers}.
\newblock In \emph{Proceedings of the 58th Annual Meeting of the Association for Computational Linguistics}, pages 975--984, Online. Association for Computational Linguistics.

\bibitem[{Mihaylov et~al.(2018)Mihaylov, Clark, Khot, and Sabharwal}]{mihaylov-etal-2018-suit}
Todor Mihaylov, Peter Clark, Tushar Khot, and Ashish Sabharwal. 2018.
\newblock \href {https://doi.org/10.18653/v1/D18-1260} {Can a suit of armor conduct electricity? a new dataset for open book question answering}.
\newblock In \emph{Proceedings of the 2018 Conference on Empirical Methods in Natural Language Processing}, pages 2381--2391, Brussels, Belgium. Association for Computational Linguistics.

\bibitem[{Murthy et~al.(2025)Murthy, Kumar, Venkateswaran, and Contractor}]{murthy2025evaluating}
Rudra Murthy, Prince Kumar, Praveen Venkateswaran, and Danish Contractor. 2025.
\newblock \href {https://openreview.net/forum?id=qit4pa6PpY} {Evaluating the instruction-following abilities of language models using knowledge tasks}.

\bibitem[{Ouyang et~al.(2022)Ouyang, Wu, Jiang, Almeida, Wainwright, Mishkin, Zhang, Agarwal, Slama, Ray et~al.}]{ouyang2022training}
Long Ouyang, Jeffrey Wu, Xu~Jiang, Diogo Almeida, Carroll Wainwright, Pamela Mishkin, Chong Zhang, Sandhini Agarwal, Katarina Slama, Alex Ray, and 1 others. 2022.
\newblock Training language models to follow instructions with human feedback.
\newblock \emph{Advances in neural information processing systems}, 35:27730--27744.

\bibitem[{Pang and Lee(2005)}]{pang2005seeing}
Bo~Pang and Lillian Lee. 2005.
\newblock Seeing stars: Exploiting class relationships for sentiment categorization with respect to rating scales.
\newblock In \emph{Proceedings of the 43rd Annual Meeting of the Association for Computational Linguistics}, pages 115--124. Association for Computational Linguistics.

\bibitem[{Qin et~al.(2024)Qin, Song, Hu, Yao, Cho, Wang, Wu, Liu, Liu, and Yu}]{qin2024infobench}
Yiwei Qin, Kaiqiang Song, Yebowen Hu, Wenlin Yao, Sangwoo Cho, Xiaoyang Wang, Xuansheng Wu, Fei Liu, Pengfei Liu, and Dong Yu. 2024.
\newblock Infobench: Evaluating instruction following ability in large language models.
\newblock In \emph{Findings of the Association for Computational Linguistics: ACL 2024}.

\bibitem[{Robinson and Wingate(2023)}]{robinson2023leveraging}
Joshua Robinson and David Wingate. 2023.
\newblock Leveraging large language models for multiple choice question answering.
\newblock In \emph{International Conference on Learning Representations}.

\bibitem[{Roy and Roth(2015)}]{roy-roth-2015-solving}
Subhro Roy and Dan Roth. 2015.
\newblock \href {https://doi.org/10.18653/v1/D15-1202} {Solving general arithmetic word problems}.
\newblock In \emph{Proceedings of the 2015 Conference on Empirical Methods in Natural Language Processing}, pages 1743--1752, Lisbon, Portugal. Association for Computational Linguistics.

\bibitem[{Sanh et~al.(2021)Sanh, Webson, Raffel, Bach, Sutawika, Alyafeai, Chaffin, Stiegler, Scao, Raja et~al.}]{sanh2021multitask}
Victor Sanh, Albert Webson, Colin Raffel, Stephen~H Bach, Lintang Sutawika, Zaid Alyafeai, Antoine Chaffin, Arnaud Stiegler, Teven~Le Scao, Arun Raja, and 1 others. 2021.
\newblock Multitask prompted training enables zero-shot task generalization.
\newblock \emph{arXiv preprint arXiv:2110.08207}.

\bibitem[{Wang et~al.(2022)Wang, Mishra, Alipoormolabashi, Kordi, Mirzaei, Naik, Ashok, Dhanasekaran, Arunkumar, Stap et~al.}]{wang2022super}
Yizhong Wang, Swaroop Mishra, Pegah Alipoormolabashi, Yeganeh Kordi, Amirreza Mirzaei, Atharva Naik, Arjun Ashok, Arut~Selvan Dhanasekaran, Anjana Arunkumar, David Stap, and 1 others. 2022.
\newblock Super-naturalinstructions: Generalization via declarative instructions on 1600+ nlp tasks.
\newblock In \emph{Proceedings of the 2022 conference on empirical methods in natural language processing}, pages 5085--5109.

\bibitem[{Webson and Pavlick(2022)}]{webson2022prompt}
Albert Webson and Ellie Pavlick. 2022.
\newblock Do prompt-based models really understand the meaning of their prompts?
\newblock In \emph{Proceedings of the 2022 conference of the north american chapter of the association for computational linguistics: Human language technologies}, pages 2300--2344.

\bibitem[{Wei et~al.(2022)Wei, Bosma, Zhao, Guu, Yu, Lester, Du, Dai, and Le}]{wei2022finetuned}
Jason Wei, Maarten Bosma, Vincent~Y. Zhao, Kelvin Guu, Adams~Wei Yu, Brian Lester, Nan Du, Andrew~M. Dai, and Quoc~V. Le. 2022.
\newblock Finetuned language models are zero-shot learners.
\newblock In \emph{International Conference on Learning Representations}.

\bibitem[{Zeng et~al.(2024)Zeng, Yu, Gao, Meng, Goyal, and Chen}]{zeng2024evaluating}
Zhiyuan Zeng, Jiatong Yu, Tianyu Gao, Yu~Meng, Tanya Goyal, and Danqi Chen. 2024.
\newblock Evaluating large language models at evaluating instruction following.
\newblock In \emph{International Conference on Learning Representations}, volume 2024, pages 40193--40219.

\bibitem[{Zhou et~al.(2023)Zhou, Lu, Mishra, Brahma, Basu, Luan, Zhou, and Hou}]{zhou2023instruction}
Jeffrey Zhou, Tianjian Lu, Swaroop Mishra, Siddhartha Brahma, Sujoy Basu, Yi~Luan, Denny Zhou, and Le~Hou. 2023.
\newblock Instruction-following evaluation for large language models.
\newblock \emph{arXiv preprint arXiv:2311.07911}.

\end{thebibliography}

\appendix
\newpage
\section{Experimental Details}
\label{app:experimental-details}

\begin{table*}[!t]
\centering
\small
\resizebox{\textwidth}{!}{%
\begin{tabular}{l|ccc|ccc|ccc}
\toprule
\multirow{2}{*}{Model}
& \multicolumn{3}{c|}{RACE-M}
& \multicolumn{3}{c|}{ARC-C}
& \multicolumn{3}{c}{OBQA-Main} \\
\cmidrule(lr){2-4}
\cmidrule(lr){5-7}
\cmidrule(lr){8-10}
& Std.~$\uparrow$ & Non-Std.~$\downarrow$ & IFFR~$\downarrow$
& Std.~$\uparrow$ & Non-Std.~$\downarrow$ & IFFR~$\downarrow$
& Std.~$\uparrow$ & Non-Std.~$\downarrow$ & IFFR~$\downarrow$ \\
\midrule
Qwen3.5-0.8B
& $70.15_{\pm 0.98}$ & $68.99_{\pm 1.41}$ & $95.72_{\pm 2.15}$
& $64.95_{\pm 0.27}$ & $56.34_{\pm 7.62}$ & $79.75_{\pm 13.85}$
& $57.47_{\pm 1.10}$ & $47.13_{\pm 9.38}$ & $70.63_{\pm 20.64}$ \\

Qwen3.5-2B
& $81.11_{\pm 0.11}$ & $80.62_{\pm 0.15}$ & $97.97_{\pm 0.57}$
& $80.29_{\pm 0.13}$ & $76.62_{\pm 2.83}$ & $92.06_{\pm 4.10}$
& $72.60_{\pm 0.20}$ & $65.00_{\pm 7.98}$ & $83.64_{\pm 13.24}$ \\

Qwen3.5-4B
& $90.62_{\pm 0.28}$ & $74.91_{\pm 11.52}$ & $80.74_{\pm 12.85}$
& $91.73_{\pm 0.27}$ & $34.65_{\pm 25.73}$ & $36.06_{\pm 28.19}$
& $86.07_{\pm 0.23}$ & $29.33_{\pm 18.15}$ & $30.80_{\pm 20.74}$ \\

Qwen3.5-9B
& $92.76_{\pm 0.07}$ & $24.56_{\pm 6.27}$ & $25.09_{\pm 6.27}$
& $94.54_{\pm 0.05}$ & $20.03_{\pm 14.84}$ & $20.06_{\pm 15.24}$
& $91.87_{\pm 0.31}$ & $16.40_{\pm 7.81}$ & $15.73_{\pm 8.55}$ \\

Qwen3.5-27B
& $95.01_{\pm 0.11}$ & $4.25_{\pm 2.62}$ & $4.15_{\pm 2.00}$
& $97.51_{\pm 0.09}$ & $0.77_{\pm 0.38}$ & $0.53_{\pm 0.35}$
& $96.87_{\pm 0.23}$ & $1.73_{\pm 0.23}$ & $1.38_{\pm 0.12}$ \\
\bottomrule
\end{tabular}%
}
\caption{MCQA results across datasets and model sizes. Std.\ and Non-Std.\ denote
standard and non-standard accuracy. Values are mean$_{\pm \mathrm{std}}$ over
prompt variants.}
\label{tab:main-results}
\end{table*}

We evaluated five Qwen3.5 checkpoints: \texttt{Qwen/Qwen3.5-0.8B},
\texttt{Qwen/Qwen3.5-2B}, \texttt{Qwen/Qwen3.5-4B}, \texttt{Qwen/Qwen3.5-9B}, and
\texttt{Qwen/Qwen3.5-27B}. For MCQA, all final experiments used the DeepInfra API
through the OpenAI-compatible chat-completions interface. For sentiment
classification and mathematical QA, smaller models were run locally in Google Colab
with Hugging Face Transformers v5.12.1, while the 9B and 27B models were evaluated
through the DeepInfra API. API runs were launched from a CPU runtime with inference
performed remotely by DeepInfra; local GPU experiments used Colab GPUs (NVIDIA T4,
L4, and A100), with most local runs on L4.

All experiments used deterministic decoding. Local Hugging Face runs loaded models
with \texttt{torch\_dtype=torch.float16} and \texttt{device\_map="auto"}, evaluated
in \texttt{eval()} mode with greedy decoding (\texttt{do\_sample=False}). API runs
used \texttt{temperature=0}; backend precision was determined by DeepInfra and not
directly controlled. For Qwen chat models we used the chat template with
\texttt{enable\_thinking=False}, and local inputs were truncated to 4096 tokens.

Predictions were obtained as described in Section~\ref{sec:pipelines}: from the
first-token logits restricted to the option tokens (A/B/C/D) for MCQA and to the
label tokens (positive/negative) for sentiment classification, and by extracting
and normalizing the final numeric value from the generated response for
mathematical QA.

Wall-clock time varied by task, model size, backend, and API availability. As representative examples, the 27B model on multi-class sentiment classification through the DeepInfra API took approximately 1 hour and 37 minutes, while the 9B model took approximately 1 hour. The 4B model on multi-class sentiment classification took approximately 52 minutes. Local MultiArith runs on an L4 GPU were substantially faster, typically taking approximately 2--4 minutes per run. Exact wall-clock time was not recorded for every dataset-model configuration.
\section{Prompt-Variant Results}
\label{app:prompt-variant-results}

Tables \ref{tab:main-results}, \ref{tab:sentiment-mean-std-results}, and
\ref{tab:math-mean-std-results} report per-dataset results across all model sizes
as mean$_{\pm \mathrm{std}}$ over the three prompt variants, complementing the
mean accuracy and IFFR trends in Figures~\ref{fig:accuracy-results}
and~\ref{fig:iffr-all}.

\begin{table*}[t!]
\centering
\small
\resizebox{\textwidth}{!}{%
\begin{tabular}{l|ccc|ccc|ccc}
\toprule
\multirow{2}{*}{Model}
& \multicolumn{3}{c|}{Multi-class Sentiment}
& \multicolumn{3}{c|}{Rotten Tomatoes}
& \multicolumn{3}{c}{FinancialPhraseBank} \\
\cmidrule(lr){2-4}
\cmidrule(lr){5-7}
\cmidrule(lr){8-10}
& Std.~$\uparrow$ & Non-Std.~$\downarrow$ & IFFR~$\downarrow$
& Std.~$\uparrow$ & Non-Std.~$\downarrow$ & IFFR~$\downarrow$
& Std.~$\uparrow$ & Non-Std.~$\downarrow$ & IFFR~$\downarrow$ \\
\midrule
Qwen3.5-0.8B
& $88.08_{\pm 1.77}$ & $89.00_{\pm 0.82}$ & $95.71_{\pm 1.84}$
& $82.46_{\pm 0.90}$ & $83.27_{\pm 1.14}$ & $93.57_{\pm 4.68}$
& $93.40_{\pm 1.34}$ & $94.58_{\pm 1.47}$ & $97.67_{\pm 2.63}$ \\

Qwen3.5-2B
& $89.80_{\pm 0.14}$ & $83.43_{\pm 3.66}$ & $89.80_{\pm 3.93}$
& $84.55_{\pm 0.80}$ & $75.33_{\pm 6.11}$ & $84.14_{\pm 7.39}$
& $96.78_{\pm 0.77}$ & $91.71_{\pm 3.57}$ & $94.06_{\pm 3.38}$ \\

Qwen3.5-4B
& $90.03_{\pm 0.32}$ & $31.89_{\pm 19.27}$ & $29.45_{\pm 23.86}$
& $89.09_{\pm 0.33}$ & $37.90_{\pm 22.70}$ & $36.83_{\pm 29.76}$
& $94.75_{\pm 0.29}$ & $32.99_{\pm 25.24}$ & $32.54_{\pm 27.40}$ \\

Qwen3.5-9B
& $89.34_{\pm 0.26}$ & $22.76_{\pm 11.85}$ & $20.04_{\pm 14.65}$
& $90.43_{\pm 0.19}$ & $48.72_{\pm 31.14}$ & $49.92_{\pm 36.89}$
& $93.91_{\pm 1.34}$ & $55.50_{\pm 45.13}$ & $57.29_{\pm 47.46}$ \\

Qwen3.5-27B
& $90.24_{\pm 0.03}$ & $9.99_{\pm 0.39}$ & $2.74_{\pm 0.66}$
& $92.03_{\pm 0.41}$ & $14.04_{\pm 4.72}$ & $8.63_{\pm 5.33}$
& $99.49_{\pm 0.00}$ & $2.71_{\pm 1.28}$ & $2.38_{\pm 1.28}$ \\
\bottomrule
\end{tabular}%
}
\caption{Sentiment classification results as mean$_{\pm \mathrm{std}}$ across the
three prompt variants.}
\label{tab:sentiment-mean-std-results}
\end{table*}
\begin{table*}[t!]
\centering
\small
\resizebox{\textwidth}{!}{%
\begin{tabular}{l|ccc|ccc|ccc}
\toprule
\multirow{2}{*}{Model}
& \multicolumn{3}{c|}{MAWPS}
& \multicolumn{3}{c|}{Calc-asdiv-a}
& \multicolumn{3}{c}{MultiArith} \\
\cmidrule(lr){2-4}
\cmidrule(lr){5-7}
\cmidrule(lr){8-10}
& Std.~$\uparrow$ & Non-Std.~$\downarrow$ & IFFR~$\downarrow$
& Std.~$\uparrow$ & Non-Std.~$\downarrow$ & IFFR~$\downarrow$
& Std.~$\uparrow$ & Non-Std.~$\downarrow$ & IFFR~$\downarrow$ \\
\midrule
Qwen3.5-0.8B
& $27.04_{\pm 2.45}$ & $27.33_{\pm 10.14}$ & $60.83_{\pm 19.60}$
& $35.36_{\pm 5.07}$ & $32.30_{\pm 10.12}$ & $60.14_{\pm 17.11}$
& $8.70_{\pm 2.57}$ & $14.81_{\pm 6.72}$ & $60.69_{\pm 27.42}$ \\

Qwen3.5-2B
& $39.81_{\pm 1.72}$ & $11.36_{\pm 3.48}$ & $24.21_{\pm 10.99}$
& $55.31_{\pm 1.48}$ & $14.86_{\pm 6.46}$ & $24.18_{\pm 11.15}$
& $5.93_{\pm 0.32}$ & $3.70_{\pm 0.32}$ & $21.51_{\pm 9.97}$ \\

Qwen3.5-4B
& $71.27_{\pm 3.25}$ & $18.59_{\pm 3.52}$ & $25.23_{\pm 5.58}$
& $88.51_{\pm 0.93}$ & $27.12_{\pm 5.57}$ & $29.73_{\pm 6.73}$
& $48.33_{\pm 9.62}$ & $11.67_{\pm 3.89}$ & $19.91_{\pm 5.11}$ \\

Qwen3.5-9B
& $80.66_{\pm 0.99}$ & $4.23_{\pm 0.49}$ & $4.19_{\pm 0.65}$
& $80.05_{\pm 21.48}$ & $4.68_{\pm 3.01}$ & $6.31_{\pm 5.62}$
& $78.33_{\pm 1.92}$ & $4.63_{\pm 0.85}$ & $5.19_{\pm 0.70}$ \\

Qwen3.5-27B
& $93.61_{\pm 0.32}$ & $5.63_{\pm 3.38}$ & $5.61_{\pm 3.30}$
& $97.10_{\pm 0.17}$ & $1.37_{\pm 1.18}$ & $1.41_{\pm 1.21}$
& $97.04_{\pm 0.32}$ & $17.22_{\pm 8.55}$ & $17.73_{\pm 8.77}$ \\
\bottomrule
\end{tabular}%
}
\caption{Mathematical QA results as mean$_{\pm \mathrm{std}}$ across the three
prompt variants.}
\label{tab:math-mean-std-results}
\end{table*}

\end{document}